\documentclass[letterpaper,10pt,conference]{ieeeconf}
\IEEEoverridecommandlockouts
\usepackage{amsmath,amssymb,bm}
\usepackage{graphicx}
\usepackage{booktabs}
\usepackage{multirow}
\usepackage{threeparttable}
\usepackage{stfloats}
\usepackage{float}
\usepackage{url}
\usepackage{xcolor}
\usepackage{cite}

\newcommand{\ours}{SHARE}

\newcommand{\vect}[1]{\bm{#1}}

\title{\ours: Sharing Human Actions across Robot Embodiments\\through Unified Retargeting and Tracking}

\author{Junzhe Xu$^{1,2,*}$, Haoyu Zhang$^{2,*}$, Yanheng Mai$^{1}$, Yunheng Wang$^{1}$, Taowen Wang$^{1}$, Lusong Li$^{2}$,\\ Liang Lin$^{2}$, Xiaodong He$^{2}$, Zecui Zeng$^{2,\dagger}$, Renjing Xu$^{1,\dagger}$
\thanks{$^{1}$ The Hong Kong University of Science and Technology (Guangzhou), Guangzhou, China.}%
\thanks{$^{2}$ JD Explore Academy, Beijing, China.}%
\thanks{$^\dagger$ Corresponding authors.}%
\thanks{$^*$ Equal contribution.}
}

\begin{document}
\maketitle

\begin{abstract}
Reproducing a human action across humanoid robots requires both motion retargeting and closed-loop whole-body tracking, yet training separate per-robot retargeters and controllers duplicates motion coordination and inflates multi-platform maintenance cost. We introduce \ours{}, pairing the cross-embodiment retargeter AdaMorph with the shared tracking policy AdaTrack. AdaMorph combines target-agnostic motion encoding with robot-conditioned decoding to synthesize native references, while AdaTrack processes variable-length joint and morphology sequences through a shared network and action head trained via expert distillation and reinforcement learning. In MuJoCo evaluations across Unitree G1, HighTorque HI, PND Adam Lite, and Booster T1, AdaTrack achieved 94.00\%/91.96\% seen/unseen success and 49.1/50.9~mm mean per-keypoint position errors using 710,120 actor parameters, 21.15\% of four specialist actors' total. At 1\% target data, cross-embodiment pretraining improved held-out-robot adaptation over scratch by 9.67/13.62 percentage points on seen/unseen motions, and squatting, dancing, and boxing on physical G1 and T1 robots demonstrate cross-platform whole-body execution.
\end{abstract}

\section{Introduction}
Motion capture records coordinated whole-body movement across locomotion, gestures, and dance, providing a rich behavioral foundation for humanoid robots~\cite{mahmood2019amass}. Reproducing a captured human action across heterogeneous humanoids, however, faces coupled kinematic and dynamic constraints. Because platforms differ in stature, limb proportions, joint counts, and actuator limits, the same human behavior must first be mapped onto distinct, feasible robot joint trajectories. During physical execution, shifting limb momentum and ground-contact transitions further perturb balance, requiring continual closed-loop feedback tailored to each body's inertia and actuation interface. Sharing human actions across embodiments therefore couples two interdependent stages: generating embodiment-compatible references and tracking them through closed-loop whole-body control.

Recent humanoid retargeting methods advance geometric scaling, interaction preservation, and learned sequence generation~\cite{araujo2025gmr,yang2025omniretarget,zhao2026nmr}, while whole-body controllers translate reference trajectories into expressive, agile hardware execution~\cite{cheng2024expressive,he2025omnih2o,liao2026beyondmimic}. In parallel, morphology-aware architectures and multi-embodiment policies address structural variation through body-graph attention and shared control representations~\cite{sferrazza2025body,bohlinger2025urma,li2026mimicall}. Despite these advances, building an independent retargeter and fixed-dimensional tracker for every robot duplicates the learning of whole-body coordination and increases maintenance overhead as platforms grow, whereas mapping diverse robots onto a padded common joint template obscures native actuator semantics. Bridging learned multi-robot reference generation with native-space shared feedback---and reusing that control experience when adapting to a held-out embodiment under scarce demonstrations---remains an open system challenge.

Our key insight is to separate transferable motion coordination from embodiment-conditioned realization across both stages. Raising an arm while shifting body weight, for example, entails coupled upper- and lower-body timing across platforms, whereas feasible joint excursions, inertia-dependent corrective responses, and joint-command dimensions depend on the target robot. Encoding the temporal ordering and rhythm of human motion independently of the target platform allows one latent sequence to be decoded under explicit robot conditioning. During execution, exposing static kinematic, inertial, and actuator descriptors alongside real-time joint states enables a single shared policy to compute embodiment-appropriate feedback directly in each robot's native action space.

\begin{figure*}[t]
 \centering
 \includegraphics[width=\textwidth]{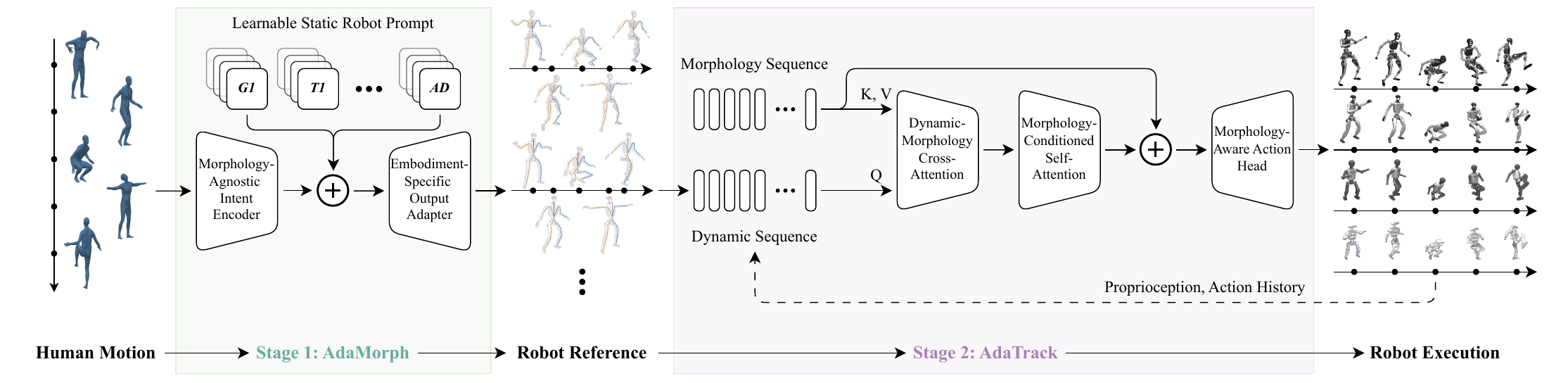}
 \caption{Two-stage architecture and data flow of \ours{}. In Stage 1, AdaMorph generates robot references through target-agnostic motion encoding, robot-conditioned decoding, and output adaptation. In Stage 2, AdaTrack forms dynamic sequences from reference commands, proprioception, and action history, combines them with morphology sequences, and predicts joint controls through a shared feedback network and action head. Reference trajectories connect retargeting to closed-loop execution.}
 \label{fig:architecture}
\end{figure*}

We propose \ours{}---\emph{\textbf{S}haring \textbf{H}uman \textbf{A}ctions across \textbf{R}obot \textbf{E}mbodiments}---unifying cross-embodiment retargeting in AdaMorph with shared closed-loop tracking in AdaTrack through 50-Hz native robot references, as illustrated in Figure~\ref{fig:architecture}. AdaMorph pairs a target-agnostic temporal motion encoder with dual-path robot-conditioned decoding and lightweight output adapters to synthesize drift-controlled reference trajectories. AdaTrack represents each robot as a variable-length token sequence over its own joints, coupling dynamic states with morphology descriptors through shared attention and one shared joint-action head. We evaluate reference fidelity, multi-embodiment closed-loop tracking, training-stage ablations, and leave-one-embodiment-out adaptation across Unitree G1 (G1), HighTorque HI (HI), PND Adam Lite (AD), and Booster T1 (T1) over 600 seen and 448 evaluation-only unseen robot--motion pairs, complemented by whole-body deployment on physical G1 and T1 humanoids.

Our contributions are threefold:
\begin{itemize}
 \item \textbf{Unified cross-robot motion generation and execution.} \ours{} connects a shared retargeter and a shared tracking policy through native 50-Hz robot references, enabling one human action to be synthesized and executed across heterogeneous platforms without a padded common joint space, validated on four simulated and two physical humanoids.
 \item \textbf{Robot-conditioned motion retargeting.} AdaMorph integrates target-agnostic motion encoding, dual-path robot-conditioned decoding, and integrated-trajectory supervision to accommodate distinct joint layouts within one sequence model, preserving whole-body activity rhythm and motion form across seen and unseen actions.
 \item \textbf{Morphology-conditioned whole-body tracking.} AdaTrack combines variable-length joint tokens, explicit physical morphology conditioning, a shared action head, and expert-initialized reinforcement learning, consolidating four specialists while providing a reusable initialization for low-data held-out adaptation.
\end{itemize}

\section{Related Work}
\textbf{Cross-Embodiment Reference Generation.} Recent retargeters link reference quality directly to downstream control. General Motion Retargeting (GMR) pairs non-uniform human-to-robot scaling with two-stage inverse kinematics to mitigate tracking-degrading artifacts~\cite{araujo2025gmr}, and OmniRetarget preserves robot--object--terrain spatial and contact relations via interaction-mesh optimization~\cite{yang2025omniretarget}. Whereas both solve per-target geometric optimization, Neural Motion Retargeting (NMR) supervises a sequence network on physics-refined rollouts for G1~\cite{zhao2026nmr}. AdaMorph instead shares one learned generator across heterogeneous humanoids via robot-conditioned decoding, output adapters, and integrated-trajectory supervision, validated under closed-loop AdaTrack control.

\textbf{Whole-Body Tracking and Cross-Embodiment Control.} Expressive humanoid controllers include ExBody for upper-body imitation with velocity-tracked locomotion~\cite{cheng2024expressive}, OmniH2O for teacher--student whole-body teleoperation and autonomy~\cite{he2025omnih2o}, and BeyondMimic for agile tracking distilled into guided diffusion~\cite{liao2026beyondmimic}. Extending control across embodiments further requires structural conditioning: Body Transformer applies masked attention over sensor--actuator graphs~\cite{sferrazza2025body}, URMA shares locomotion via morphology-agnostic encoders and decoders~\cite{bohlinger2025urma}, and MimicAll conditions whole-body tracking on structured embodiment descriptors over a common semantic joint interface~\cite{li2026mimicall}. AdaTrack instead operates on native variable-length joint sequences without a padded semantic joint space, coupling physical descriptors with one shared joint-action head trained via expert distillation and reinforcement learning.

\section{\ours{}: System Overview}
As shown in Figure~\ref{fig:architecture}, \ours{} decouples cross-embodiment action transfer into open-loop reference generation in AdaMorph and closed-loop feedback tracking in AdaTrack. Rather than projecting heterogeneous humanoids into a synthetic common joint space, both stages share internal sequence and feedback computation while keeping trajectory and actuator interfaces native to each robot.

Let $e\in\mathcal{E}$ denote a configured embodiment with $J_e$ controlled joints and native reference space $\mathcal{K}_e$. Given a human motion sequence $\vect{X}^{h}=\{\vect{x}^{h}_{t}\}_{t=1}^{T}$ and body shape $\vect{\beta}$, AdaMorph predicts a 50-Hz reference trajectory:
\begin{equation}
 \hat{\vect{Y}}^e=f_{\phi}(\vect{X}^{h},\vect{\beta},e)\in\mathcal{K}_e^T.
\end{equation}
At control step $t$, AdaTrack receives proprioception $\vect{o}_t^e$, reference command $\vect{r}_t^e$ extracted from $\hat{\vect{Y}}^e$, previous action $\vect{a}_{t-1}^e$, and static morphology descriptors $\mathcal{M}_e$, mapping them through one shared actor to a native action:
\begin{equation}
 \vect{a}_t^e=\pi_{\theta}(\vect{o}_t^e,\vect{r}_t^e,
 \vect{a}_{t-1}^e,\mathcal{M}_e)\in\mathbb{R}^{J_e}.
\end{equation}
Root motion and native joint trajectories in $\hat{\vect{Y}}^e$ form the interface between stages: AdaMorph resolves kinematic adaptation from human motion to each robot's joint layout via shared parameters $\phi$, while AdaTrack uses a single parameter set $\theta$ to extract real-time tracking errors and stabilize physical interaction across embodiments.

\section{AdaMorph: Cross-Embodiment Reference Generation}
\label{sec:adamorph}
Directly mapping human poses to robot joints in a single robot-specific network entangles source behavioral rhythm with target joint topology. In Stage~1 of Figure~\ref{fig:architecture}, AdaMorph instead encodes temporal human motion once and conditions trajectory decoding on the target robot, sharing the Transformer encoder--decoder backbone while isolating topology-dependent dimensions to robot prompts and output adapters:
\begin{align}
 \vect{P}_h &= A_h(\vect{\beta}),\nonumber\\
 \vect{Z} &= \operatorname{Enc}([\vect{P}_h;E_h(\vect{X}^h)])_{\rm motion},\nonumber\\
 \vect{H}^e &= \operatorname{Dec}(\vect{Z},\vect{P}_e),\nonumber\\
 \hat{\vect{Y}}^e &= \psi_e(\vect{H}^e),
\end{align}
where $A_h$ constructs the Dynamic Human Prompt, $E_h$ projects per-frame human motion features, and $[\cdot;\cdot]$ concatenates token sequences. The shared encoder returns frame-aligned motion latents $\vect{Z}$, the shared decoder conditions them on a Learnable Static Robot Prompt $\vect{P}_e$, and the Embodiment-Specific Output Adapter $\psi_e$ maps decoded features $\vect{H}^e$ to native root and joint channels.

\subsection{Canonical Motion Interface}
Built from pose and shape parameters of the SMPL human model family~\cite{loper2015smpl}, the Canonical Motion Interface standardizes root-motion semantics across bodies without requiring explicit human--robot joint pairing inside the network. Each human frame is parameterized as
\begin{equation}
 \vect{x}^{h}_{t}=
 [\vect{v}_{t},\vect{\omega}_{t},\vect{g}_{t},\vect{J}_{t},z_t],
 \qquad
 \vect{g}_{t}=\vect{R}_{t}^{\top}[0,0,-1]^{\top},
\end{equation}
where $\vect{R}_t$ is root orientation, $\vect{v}_t,\vect{\omega}_t\in\mathbb{R}^3$ are local root linear and angular velocities, $\vect{g}_t\in\mathbb{R}^3$ is root-frame projected gravity, $\vect{J}_t$ concatenates continuous 6-D local joint rotations, and $z_t$ is world-frame root height. Correspondingly, each target robot frame is represented as
\begin{equation}
 \hat{\vect{y}}^{e}_{t}=
 [\hat{\vect{v}}_{t},\hat{\vect{\omega}}_{t},
  \hat{\vect{g}}_{t},\hat z_t,\hat{\vect{q}}^{e}_{t}],
 \qquad
 \hat{\vect{q}}^{e}_{t}\in\mathbb{R}^{J_e}.
\end{equation}
Expressing root velocities and gravity in the local root frame removes dependence on global horizontal capture origin and initial heading, while continuous 6-D rotations eliminate regression discontinuities~\cite{zhou2019continuity}. Because the root channels retain identical physical meaning across all robots, only the joint-position vector $\hat{\vect{q}}^{e}_{t}$ varies in dimension with $J_e$. Global orientation and horizontal displacement are reconstructed by integrating $\hat{\vect{\omega}}_t$ and $\hat{\vect{v}}_t$, whereas $\hat z_t$ is predicted directly rather than integrated so vertical motion remains anchored to the ground without accumulating vertical velocity drift.

\subsection{Morphology-Agnostic Intent Encoder}
Injecting target-robot identity into the encoder would bind latent features to a single embodiment and prevent latent reuse. The Morphology-Agnostic Intent Encoder therefore processes exclusively source-human inputs. A multilayer perceptron (MLP) maps $\vect{\beta}$ to $\vect{P}_h$, giving the encoder anthropometric context so identical joint rotations from subjects with different stature and limb proportions are interpreted consistently. The shared Transformer encodes $[\vect{P}_h;E_h(\vect{X}^h)]$ through temporal self-attention, relating each frame to surrounding motion phase and rhythm, after which the prompt positions are stripped to yield frame-aligned latents $\vect{Z}$ reusable across all target robots.

\subsection{Dual-Path Robot-Conditioned Decoding}
Translating $\vect{Z}$ into kinematically feasible trajectories requires exposing target embodiment constraints throughout decoding without duplicating the temporal backbone. The shared decoder conditions $\vect{Z}$ on a Learnable Static Robot Prompt $\vect{P}_e\in\mathbb{R}^{L_p\times d}$ of length $L_p$ and latent width $d$ via two complementary pathways. First, all $L_p$ prompt tokens serve as cross-attention keys and values. Second, their pooled mean $\vect{c}_e=L_p^{-1}\sum_i\vect{P}_{e,i}$ modulates self-attention, cross-attention, and feed-forward sublayers through feature modulation~\cite{perez2018film} and zero-initialized adaptive layer normalization (AdaLN)~\cite{peebles2023scalable}:
\begin{equation}
 \operatorname{AdaLN}_l(\vect{h},\vect{c}_e)=
 (1+\vect{\gamma}_l(\vect{c}_e))\odot\operatorname{LN}(\vect{h})+
 \vect{b}_l(\vect{c}_e),
\end{equation}
where $\operatorname{LN}$ denotes layer normalization and $\odot$ element-wise multiplication. Zero-initializing the scale and shift heads $\vect{\gamma}_l,\vect{b}_l$ reduces each sublayer to standard normalization at the start of training, stabilizing optimization of the shared decoder before embodiment specialization emerges. Cross-attention allows individual motion frames to query fine-grained components of $\vect{P}_e$ as movement phases change, whereas pooled AdaLN maintains a consistent global target context across layers; all robot-specific parameters remain confined to $\vect{P}_e$ and $\psi_e$.

\subsection{Trajectory-Consistency Objective}
Supervising per-frame velocities alone leaves long-horizon trajectory drift unconstrained: a small persistent bias in angular velocity $\hat{\vect{\omega}}_t$ progressively rotates the orientation used to integrate subsequent local linear velocities $\hat{\vect{v}}_t$, compounding into substantial horizontal path deviation. To make these coupled errors directly visible to gradient updates during training, AdaMorph initializes $\hat{\vect{R}}_0=\vect{R}_0$ and $\hat{\vect{p}}^{xy}_0=\vect{p}^{xy}_0$ from ground truth and integrates predicted velocities:
\begin{align}
 \hat{\vect{R}}_t
 &=\hat{\vect{R}}_{t-1}
 \operatorname{Exp}\!\left([\hat{\vect{\omega}}_t]_{\times}\Delta t\right),\nonumber\\
 \hat{\vect{p}}^{xy}_t
 &=\hat{\vect{p}}^{xy}_{t-1}+
 \left(\hat{\vect{R}}_t\hat{\vect{v}}_t\right)_{xy}\Delta t,
\end{align}
where $\Delta t$ is the 50-Hz frame interval, $[\cdot]_{\times}$ is the skew-symmetric operator, and $\operatorname{Exp}$ is the rotation exponential. Under backward differencing, $\hat{\vect{\omega}}_t$ represents angular motion from frame $t-1$ to $t$, with predicted velocities unnormalized to physical units prior to integration. Although $\operatorname{Exp}$ preserves $SO(3)$ in exact arithmetic, Gram--Schmidt orthogonalization is applied to correct floating-point drift in accumulated rotations and maintain orthogonality. Joint supervision minimizes
\begin{equation}
 \mathcal{L}_{\rm AdaMorph}=
 \mathcal{L}_{\rm inst}+
 \lambda_{\rm rot}(s)\mathcal{L}_{\rm rot}+
 \lambda_{\rm traj}(s)\mathcal{L}^{xy}_{\rm traj}+
 \lambda_z\mathcal{L}_{z},
\end{equation}
penalizing instantaneous frame reconstruction $\mathcal{L}_{\rm inst}$ alongside integrated orientation $\mathcal{L}_{\rm rot}$, horizontal trajectory $\mathcal{L}^{xy}_{\rm traj}$, and direct height $\mathcal{L}_{z}$ errors. Because noisy early velocity predictions would otherwise generate destabilizing long-horizon gradients, $\lambda_{\rm rot}(s)$ and $\lambda_{\rm traj}(s)$ ramp up over training step $s$, while $\mathcal{L}^{xy}_{\rm traj}$ transitions from ground-truth to predicted orientation when integrating $\hat{\vect{v}}_t$.

\section{AdaTrack: Cross-Embodiment Shared Feedback Control}
\label{sec:adatrack}
Flattening heterogeneous robots into a fixed-width observation and action vector introduces artificial padding and ties policy weights to a prescribed joint ordering. In Stage~2 of Figure~\ref{fig:architecture}, AdaTrack instead expresses joint-count variation through sequence length: dynamic and morphology encoders map real-time robot state and static structure into aligned token sequences $\vect{D}_t^e,\vect{M}^e\in\mathbb{R}^{(J_e+1)\times d}$ of width $d$, processed by a shared Transformer and decoded by one shared Morphology-Aware Action Head.

\subsection{Variable-Length Dynamic Representation}
Each robot's dynamic sequence consists of one Base Token and $J_e$ Joint Tokens. Relative to a configured torso or base tracking anchor link, the Base Token summarizes whole-body state and pose tracking error:
\begin{equation}
 \vect{b}_t^e=
 [\delta\vect{p}_t^e,\operatorname{rot6}(\delta\vect{R}_t^e),
  \vect{v}_t^{b,e},\vect{\omega}_t^{b,e}]\in\mathbb{R}^{15},
\end{equation}
combining 3-D anchor-frame relative position $\delta\vect{p}_t^e$, 6-D relative orientation $\operatorname{rot6}(\delta\vect{R}_t^e)$ formed from the first two rotation-matrix columns, and 3-D base-frame linear and angular velocities $\vect{v}_t^{b,e},\vect{\omega}_t^{b,e}$ to supply global balance context. Each Joint Token $j$ co-locates local tracking targets, measured state, and one-step control history:
\begin{equation}
 \vect{u}_{t,j}^{e}=
 [q^{\rm ref}_{t,j},\dot q^{\rm ref}_{t,j},
  q^{\rm rel}_{t,j},\dot q_{t,j},a_{t-1,j}],
\end{equation}
containing reference position and velocity $q^{\rm ref},\dot q^{\rm ref}$, position relative to the robot's default standing posture $q^{\rm rel}$, measured velocity $\dot q$, and previous action $a_{t-1,j}$. Robot-specific base and per-joint normalizers standardize channel scales across diverse actuators before separate linear projections embed base and joint tokens to width $d$, allowing token count $J_e+1$ to vary across robots with shared weights.

\subsection{Dynamic--Morphology Alignment}
An identical joint-angle or velocity tracking error demands different corrective actions depending on link mass, lever arm, joint limits, and actuator stiffness. To expose these physical differences explicitly, each dynamic token is paired with a static morphology descriptor organized into seven groups:
\begin{itemize}
 \item \textbf{element identity and control validity:} base/joint indicator and actuator validity flag;
 \item \textbf{embodiment scale:} joint count, action dimension, body count, and controllable-joint ratio;
 \item \textbf{topological position:} joint index, tree depth, and adjacency to root and neighboring joints;
 \item \textbf{joint kinematics:} rotation axis, range of motion, and hard limits;
 \item \textbf{rigid-body properties:} associated link mass, inertia, and geometric scale;
 \item \textbf{actuator properties:} effort limit, stiffness, damping, and action scale; and
 \item \textbf{semantic attributes:} anatomical region, symmetry side, and joint function.
\end{itemize}
For all four evaluated robots, joint count and action dimension both equal $J_e$ with a controllable-joint ratio of one, while the validity flag is zero for the Base Token and one for every Joint Token. A compact, joint-aligned descriptor subset is embedded into $\vect{M}^e$ for attention, whereas the full descriptor is retained for action decoding. Dynamic--Morphology Cross-Attention uses $\vect{D}_t^e$ as queries and $\vect{M}^e$ as keys and values:
\begin{equation}
 \tilde{\vect{D}}_t^e=\operatorname{LN}\!\left(
 \vect{D}_t^e+\alpha\operatorname{MHA}
 (\vect{D}_t^e+\vect{P},\vect{M}^e+\vect{P},\vect{M}^e+\vect{P})
 \right),
\end{equation}
with multi-head attention $\operatorname{MHA}$, shared positional encoding $\vect{P}$, and learned residual gate $\alpha$. Positional alignment associates each dynamic token with its own structural location, while cross-attention simultaneously allows a joint to query structural properties elsewhere in the kinematic chain before whole-body interaction.

\subsection{Morphology-Conditioned Whole-Body Feedback}
To capture state-dependent coordination across joints---such as stance-leg and trunk compensation during rapid arm swings---AdaTrack applies $L$ layers of Morphology-Conditioned Self-Attention over $\vect{H}^{0}=\tilde{\vect{D}}_t^e$, omitting $t$ and $e$ below for brevity. At layer $l$, $g_l(\vect{M}^e)$ predicts token-wise scale and shift vectors for the attention and feed-forward network (FFN) sublayers, indexed by $a$ and $f$:
\begin{align}
 \bar{\vect{H}}^{l}
 &= (1+\vect{\gamma}^{l}_{a})\odot
    \operatorname{LN}(\vect{H}^{l-1})+\vect{b}^{l}_{a},\nonumber\\
 \vect{H}^{l\prime}
 &= \vect{H}^{l-1}+\operatorname{MHA}(\bar{\vect{H}}^{l}),\nonumber\\
 \vect{H}^{l}
 &= \vect{H}^{l\prime}+\operatorname{FFN}\!\left(
    (1+\vect{\gamma}^{l}_{f})\odot
    \operatorname{LN}(\vect{H}^{l\prime})+\vect{b}^{l}_{f}
    \right).
\end{align}
Cross-attention grounds tokens in structural context, token-wise AdaLN preserves static morphology across layers, and self-attention couples joints by instantaneous pose error and support state, separating rollout-invariant structure from dynamic coordination.

\subsection{Shared Joint-Action Decoding}
Rather than forcing the Transformer hidden state to compress all actuator limits and scaling parameters, the shared Morphology-Aware Action Head concatenates joint $j$'s final dynamic feature $\vect{h}_{t,j}^{L}$ directly with its full morphology descriptor $\vect{c}_{j}^{e}$ to predict a scalar action mean:
\begin{equation}
 \mu_{t,j}^{e}=g_{\theta}\!\left([\vect{h}_{t,j}^{L};\vect{c}_{j}^{e}]\right).
\end{equation}
The shared head processes only the $J_e$ Joint Tokens, excluding the Base Token, and assembles outputs in native joint order into $\vect{\mu}_t^e\in\mathbb{R}^{J_e}$. At deployment, deterministic actions $\vect{a}_t^e=\vect{\mu}_t^e$ map to joint-position targets via each robot's action scales and default posture, discarding training-only critics and exploration noise.

\subsection{Three-Stage Curriculum for Shared Feedback Learning}
Optimizing a multi-robot policy from scratch via reinforcement learning (RL) entangles native action-space grounding with closed-loop balance stabilization across heterogeneous bodies. We therefore decouple training into specialist learning, multi-robot distillation, and closed-loop RL, keeping AdaMorph frozen with no gradient flow between stages.

\textbf{Expert training and demonstration collection.} We first train an independent native-space MLP expert $\pi_{E,e}$ for each robot via proximal policy optimization (PPO)~\cite{schulman2017ppo} on frozen AdaMorph references. Rolling out each expert yields native-space demonstrations:
\begin{align}
 \mathcal{D}_e
 &=\{(\vect{s}_i^e,\vect{a}_i^{E,e})\}_{i=1}^{N_e},\nonumber\\
 \vect{s}_i^e
 &=(\vect{o}_i^e,\vect{r}_i^e,\vect{a}_{\mathrm{prev},i}^{E,e},\mathcal{M}_e),\nonumber\\
 \vect{a}_i^{E,e}
 &=\pi_{E,e}(\vect{o}_i^e,\vect{r}_i^e,\vect{a}_{\mathrm{prev},i}^{E,e}),
\end{align}
where $N_e$ is the transition count, $i$ indexes dataset samples rather than time steps, and $\vect{a}_{\mathrm{prev},i}^{E,e}$ stores the preceding executed expert action within the same rollout, using the reset value at episode starts. Solving single-robot stabilization first provides clean, native-dimensional action targets for cross-embodiment consolidation.

\textbf{Multi-robot expert distillation.} Behavior cloning (BC) distills $\mathcal{D}=\bigcup_e\mathcal{D}_e$ into the shared actor:
\begin{equation}
 \mathcal{L}_{\rm BC}=
 \mathbb{E}_{e,(\vect{s},\vect{a}^{E})\sim\mathcal{D}_e}
 \left[\operatorname{SmoothL1}
 (\vect{\mu}_{\theta}(\vect{s}),\vect{a}^{E})\right].
\end{equation}
Constructing each mini-batch from a single embodiment preserves native token lengths without padding while accumulating gradients across all robots into $\theta$, grounding the multi-robot action mapping on expert states.

\textbf{Closed-loop reinforcement learning.} Because offline imitation does not expose the shared policy to compounding rollout errors, the distilled actor next interacts closed-loop with all robot environments on frozen AdaMorph references. PPO optimizes pose and velocity tracking rewards with smoothness, limit, and collision penalties:
\begin{equation}
 \pi_{\theta}^{e}(\vect{a}_t^e\mid\vect{s}_t^e)=
 \mathcal{N}\!\left(
 \vect{\mu}_{\theta}(\vect{s}_t^e),
 \operatorname{diag}((\vect{\sigma}^{e})^2)\right),
\end{equation}
pairing shared mean $\vect{\mu}_{\theta}$ with robot-specific exploration scales $\vect{\sigma}^{e}$ and privileged per-robot asymmetric critics $V_{\eta_e}(\vect{o}_{V,t}^e)$. Critics warm up with $\theta$ frozen before clipped PPO~\cite{schulman2017ppo} updates actors and critics under fixed distillation normalizers.

\section{Experiments}
\subsection{Experimental Setup and Evaluation Protocol}
We evaluate \ours{} on G1, HI, AD, and T1 across reference fidelity, multi-embodiment tracking, curriculum ablation, the accuracy--parameter trade-off, held-out-embodiment adaptation, and physical G1 and T1 deployment.

\begin{figure}[t]
 \centering
 \includegraphics[width=.96\columnwidth]{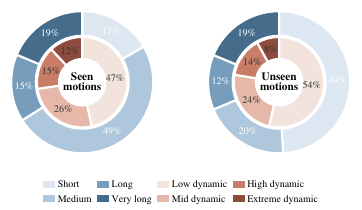}
 \caption{Source-motion distributions by duration (outer rings) and composite-dynamics percentile (inner rings); percentages are within-partition.}
 \label{fig:motion_distributions}
\end{figure}

Closed-loop MuJoCo evaluations use 50-Hz AdaMorph references under identical splits and metrics across robots. The \emph{seen} partition comprises 150 motions and 439,675 source frames from HumanEva~\cite{sigal2010humaneva}, LaFAN1~\cite{harvey2020lafan}, MoSh~\cite{loper2014mosh}, PosePrior~\cite{akhter2015pose}, the SFU Motion Capture Database~\cite{sfu_mocap}, SOMA~\cite{ghorbani2021soma}, and TotalCapture~\cite{trumble2017totalcapture} for training and in-distribution testing; the \emph{unseen} partition comprises 112 motions and 523,045 source frames from CMU~\cite{cmu_mocap}, KIT~\cite{mandery2016kit}, and DanceDB~\cite{stavrakis2012dancedb}, withheld from all training stages, yielding 600 seen and 448 unseen robot--motion pairs. Figure~\ref{fig:motion_distributions} groups clip duration into $<10$~s, $10$--$30$~s, $30$--$60$~s, and $\geq60$~s, with a composite-dynamics score averaging corpus percentile ranks of 95th-percentile root speed, 95th-percentile root acceleration, vertical root excursion, and 95th-percentile frame-to-frame SMPL-X pose change.

\textbf{Training settings.} BC uses 120,000 updates with a global batch size of 32,768. PPO refinement uses 8,192 environments per robot, 24-step rollouts, and 15,000 iterations, including a 500-iteration critic warmup. Actor normalizers remain fixed during PPO to retain the input scaling established by distillation.

\textbf{Evaluation metrics.} Closed-loop success rate is the fraction of clips completed without falling; completion is the fraction of each clip traversed. Tracking precision is measured by mean per-keypoint position error (MPKPE) and root-relative MPKPE (R-MPKPE). Evaluation is deterministic, with one rollout per robot--motion pair; all methods use the same motion sets and termination conditions.

\subsection{AdaMorph Retargeting Performance}
\begin{figure}[t]
 \centering
 \includegraphics[width=.96\columnwidth]{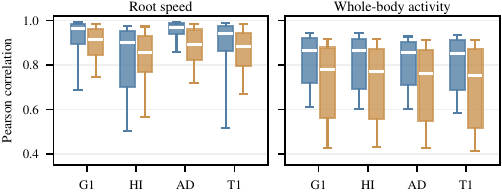}
 \caption{AdaMorph semantic consistency across 1,048 robot--motion pairs. Blue/left denotes seen and orange/right unseen motions; boxes show interquartile ranges, white lines medians, and whiskers the 10th--90th percentiles.}
 \label{fig:adamorph_consistency}
\end{figure}

\begin{table*}[t]
\caption{AdaMorph retargeting results: medians over robot--motion pairs. PCC measures semantic consistency; velocity RMSE and orientation/joint MAE measure seen-label fidelity.}
\vspace{-0.2cm}
\label{tab:adamorph_quantitative}
\centering
\begin{tabular}{cccccccc}
\hline
\multirow{2}{*}{Robot} & \multicolumn{2}{c}{Seen semantic consistency} & \multicolumn{2}{c}{Unseen semantic consistency} & \multicolumn{3}{c}{Seen label fidelity}\\
\cline{2-8}
& Root PCC & Activity PCC & Root PCC & Activity PCC & Root vel. (m/s) & Root ori. ($^\circ$) & Joint ($^\circ$)\\
\hline
G1 & 0.965 & 0.864 & 0.913 & 0.779 & 0.269 & 8.13 & 0.426\\
HI & 0.900 & 0.866 & 0.856 & 0.770 & 0.310 & 9.28 & 0.388\\
AD & 0.969 & 0.855 & 0.895 & 0.761 & 0.575 & 9.68 & 0.383\\
T1 & 0.943 & 0.854 & 0.882 & 0.753 & 0.382 & 9.45 & 0.387\\
\hline
All & 0.951 & 0.860 & 0.894 & 0.763 & 0.347 & 9.11 & 0.399\\
\hline
\end{tabular}
\end{table*}

\begin{figure}[t]
 \centering
 \includegraphics[width=\columnwidth]{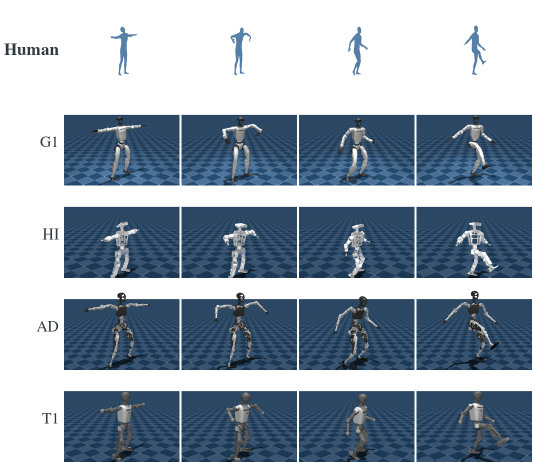}
 \caption{AdaMorph realization of one unseen action at four aligned phases. Rows show the human source and G1, HI, AD, and T1 references under fixed per-row camera scale.}
 \label{fig:adamorphqual}
\end{figure}

AdaMorph is supervised with target labels from \texttt{robot\_retargeter}~\cite{ccrpRobotRetargeter} and evaluated on temporal rhythm preservation and seen-label reconstruction. For root positions $\vect{p}_t^k$, $k\in\{h,e\}$, speed is
\begin{equation}
 s_t^k=\frac{\|\vect{p}_{t+1}^k-\vect{p}_t^k\|_2}{\Delta t_k}.
\end{equation}
Whole-body activity averages joint angular speed without requiring joint-to-joint correspondence:
\begin{align}
 A_t^h&=\frac{1}{J_h\Delta t_h}\sum_{j=1}^{J_h}
 \operatorname{angle}\!\left((\vect{R}_{t,j}^h)^\top\vect{R}_{t+1,j}^h\right),\nonumber\\
 A_t^e&=\frac{1}{J_e\Delta t_e}\sum_{j=1}^{J_e}|q_{t+1,j}^e-q_{t,j}^e|,
\end{align}
where $J_h$ is the human body-joint count, $\vect{R}_{t,j}^h$ is local joint rotation, $\operatorname{angle}(\cdot)$ returns inter-frame rotation in radians, and $\Delta t_h,\Delta t_e$ are sampling intervals. Interpolating human profiles to robot timestamps, we compute the Pearson correlation coefficient (PCC) with temporal means $\bar x,\bar y$:
\begin{equation}
 \rho(x,y)=\frac{\sum_t(x_t-\bar x)(y_t-\bar y)}
 {\sqrt{\sum_t(x_t-\bar x)^2\sum_t(y_t-\bar y)^2}},
\end{equation}
defining median-summarized per-pair consistency metrics:
\begin{equation}
 C_{\rm root}=\rho(s^h,s^e),\qquad
 C_{\rm activity}=\rho(A^h,A^e).
\end{equation}
On seen motions, we also report root-velocity root mean square error (RMSE), initial-orientation-aligned 3-D root-rotation angle error, and joint-position mean absolute error (MAE) in Table~\ref{tab:adamorph_quantitative} and Figure~\ref{fig:adamorph_consistency}.

AdaMorph achieved median activity PCCs of 0.860/0.763 and root-speed PCCs of 0.951/0.894 on seen/unseen motions, alongside seen joint and root-orientation MAEs of $0.399^\circ$ and $9.11^\circ$. Qualitatively, Figure~\ref{fig:adamorphqual} shows all four embodiments preserving the progression from arm extension to leg lift via proportion-adapted poses, turning one human clip into continuous multi-robot references.

\subsection{AdaTrack Cross-Embodiment Tracking Performance}
\begin{table}[t]
\caption{Per-embodiment AdaTrack results. Success is a percentage; MPKPE is in millimeters.}
\vspace{-0.2cm}
\label{tab:perrobot}
\centering
\begin{tabular}{ccccc}
\hline
\multirow{2}{*}{Robot} & \multicolumn{2}{c}{Seen} & \multicolumn{2}{c}{Unseen}\\
\cline{2-5}
& Success & MPKPE & Success & MPKPE\\
\hline
G1 & 90.67 & 50.5 & 89.29 & 55.7\\
HI & 92.67 & \textbf{37.3} & 90.18 & \textbf{35.5}\\
AD & 96.00 & 65.2 & 91.96 & 69.3\\
T1 & \textbf{96.67} & 43.3 & \textbf{96.43} & 43.0\\
\hline
All & 94.00 & 49.1 & 91.96 & 50.9\\
\hline
\end{tabular}
\end{table}

\begin{table*}[t]
\caption{AdaTrack training-method ablation and specialist comparison. Success and completion are percentages; position errors are millimeters.}
\vspace{-0.2cm}
\label{tab:learning_routes}
\centering
\setlength{\tabcolsep}{4.2pt}
\begin{tabular}{lcccccccc}
\hline
\multirow{2}{*}{Controller} & \multicolumn{4}{c}{Seen} & \multicolumn{4}{c}{Unseen}\\
\cline{2-9}
& Success & Completion & MPKPE & R-MPKPE & Success & Completion & MPKPE & R-MPKPE\\
\hline
Robot-specific & \textbf{97.67} & \textbf{99.21} & \textbf{34.3} & \textbf{29.6} & 91.29 & 96.08 & \textbf{42.5} & \textbf{39.3}\\
AdaTrack BC-only & 56.17 & 70.87 & 51.6 & 44.9 & 56.25 & 72.69 & 53.3 & 48.1\\
AdaTrack PPO-only & 46.83 & 64.42 & 72.4 & 69.8 & 53.13 & 69.27 & 69.3 & 64.8\\
AdaTrack BC+PPO & 94.00 & 97.19 & 49.1 & 43.5 & \textbf{91.96} & \textbf{96.22} & 50.9 & 46.4\\
\hline
\end{tabular}
\end{table*}

\begin{figure}[t]
 \centering
 \includegraphics[width=\columnwidth]{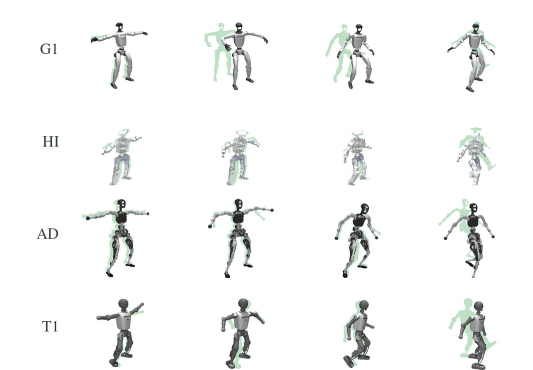}
 \caption{AdaTrack closed-loop execution at the four phases in Figure~\ref{fig:adamorphqual}. Opaque robots show actual shared-actor rollouts; translucent green silhouettes show target references. Rows are G1, HI, AD, and T1.}
 \label{fig:adatrack_demo}
\end{figure}

In Table~\ref{tab:perrobot}, the shared AdaTrack actor achieved 90.67--96.67\% seen and 89.29--96.43\% unseen success across the four robots, averaging 94.00\%/91.96\% success and 49.1/50.9~mm MPKPE. T1 reached 96.43\% unseen success, HI obtained 35.5~mm unseen MPKPE, and AD sustained 91.96\% unseen success despite a 69.3~mm offset on its larger frame. Figure~\ref{fig:adatrack_demo} overlays rollouts against the unseen references of Figure~\ref{fig:adamorphqual}, confirming multi-topology tracking with one policy.

\subsection{Training-Method Ablation}
Table~\ref{tab:learning_routes} compares BC-only, PPO-only, and BC+PPO under identical architecture and evaluation settings to isolate expert initialization and closed-loop refinement.

BC+PPO reached 94.00\% seen and 91.96\% unseen success with over 96\% completion on both splits, surpassing 56.17\% seen success for BC-only and 46.83\% for PPO-only. Compared with BC-only, PPO refinement raised seen success by 37.83 percentage points while reducing MPKPE from 51.6 to 49.1~mm, indicating that closed-loop RL primarily drives sustained trajectory execution while expert distillation grounds the initial action mapping. Seen R-MPKPE likewise dropped from 69.8~mm under PPO-only to 43.5~mm under BC+PPO, confirming improved relative body configuration alongside higher completion.

\subsection{Accuracy--Scale Trade-off of Shared Control}
The shared actor uses 710,120 parameters, a 78.85\% reduction from the four robot-specific MLP specialists' combined 3,357,132 in Table~\ref{tab:learning_routes}, consolidating cross-robot tracking into one compact policy.

Specialists achieved higher seen success at 97.67\% versus 94.00\% and lower seen/unseen MPKPE at 34.3/42.5~mm versus 49.1/50.9~mm, whereas the shared actor reached 91.96\% unseen success compared with 91.29\% for specialists. Parameter sharing thus trades specialist precision and seen-motion margin for one compact network that preserves unseen completion and supplies reusable adaptation weights.

\subsection{Cross-Embodiment Transfer of Shared Feedback}
\begin{table*}[t]
\centering
\begin{threeparttable}
\caption{Per-embodiment success-rate gains from cross-embodiment pretraining over training from scratch at each target-data fraction.}
\vspace{-0.2cm}
\label{tab:perfold_transfer}
\begin{tabular}{cccccc}
\hline
Held-out embodiment & 1\% & 5\% & 10\% & 25\% & 100\%\\
\hline
G1 & $+6.67/+8.93$ & $+9.33/+6.25$ & $+10.00/+1.79$ & $+10.00/+14.29$ & $+3.33/+3.57$\\
HI & $+14.00/+11.61$ & $+7.33/+17.86$ & $+14.00/+9.82$ & $+8.67/+6.25$ & $+2.00/+0.00$\\
AD & $+6.67/+13.39$ & $+4.67/-0.89$ & $+7.33/+1.79$ & $+0.67/+6.25$ & $+0.67/+1.79$\\
T1 & $+11.33/+20.54$ & $+14.67/+16.07$ & $+2.67/+0.89$ & $+1.33/+5.36$ & $+2.00/+2.68$\\
\hline
Aggregate & $\mathbf{+9.67/+13.62}$ & $\mathbf{+9.00/+9.82}$ & $\mathbf{+8.50/+3.57}$ & $\mathbf{+5.17/+8.04}$ & $\mathbf{+2.00/+2.01}$\\
\hline
\end{tabular}
\begin{tablenotes}[flushleft]
\scriptsize
\item[] Each entry is the seen/unseen success-rate gain in percentage points.
\end{tablenotes}
\end{threeparttable}
\end{table*}

\begin{figure}[t]
 \centering
 \includegraphics[width=.96\columnwidth]{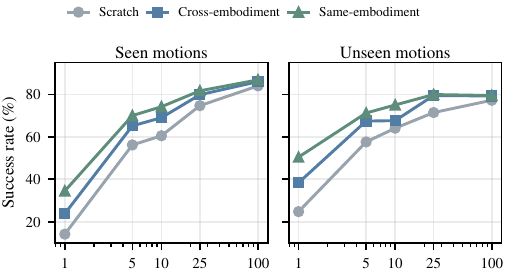}
 \caption{Held-out-embodiment success rate versus target-data fraction (\%), aggregated over four folds. Cross-embodiment pretraining improves on scratch; same-embodiment pretraining leads, with the largest gap at low budgets.}
 \label{fig:fewshot}
\end{figure}

In Figure~\ref{fig:fewshot} and Table~\ref{tab:perfold_transfer}, we hold out G1, HI, AD, and T1 in turn and adapt on 1\%--100\% target data, comparing cross-embodiment pretraining on the other three robots against scratch and same-embodiment pretraining on the target robot. Transferring only shared actor weights and refitting target normalizers, all initializations run 120,000 offline BC updates on identical nested, source-stratified demonstrations.

Cross-embodiment pretraining improved seen/unseen success over scratch by 9.67/13.62 percentage points at 1\% data, rising from 14.17\%/24.78\% to 23.83\%/38.39\%, and by 9.00/9.82 percentage points at 5\%. Seen motions excluded from adaptation gained 9.80 percentage points at 1\%, confirming transfer beyond adaptation clips, while unseen gains at 1\% spanned all four robots from 8.93 percentage points on G1 to 20.54 percentage points on T1.

Gains narrowed to 2.00/2.01 percentage points at 100\% data, concentrating the benefit in the low-data regime. Same-embodiment pretraining remained strongest, while cross-embodiment transfer varied across robots, including an unseen decrease of 0.89 percentage points for AD at 5\% data, establishing shared feedback as a reusable low-data prior.

\subsection{Real-Robot Deployment}
\begin{figure}[t]
 \centering
 \includegraphics[width=\columnwidth]{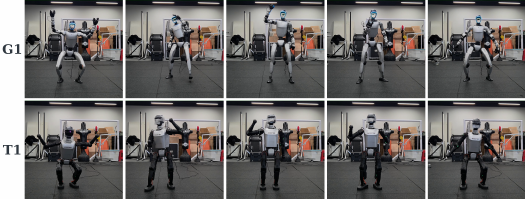}
 \caption{Physical execution on G1 (top) and T1 (bottom). Each panel shows a snapshot from a different motion recording.}
 \label{fig:hardware}
\end{figure}

We deployed \ours{} on physical G1 and T1 humanoids with 29 and 23 controlled joints using AdaMorph references and the 50-Hz shared AdaTrack actor exported in Open Neural Network Exchange (ONNX) format. Lightweight runtime adapters handled robot-specific normalization, joint ordering, default postures, and action scales for low-level position tracking. Across squatting, dancing, and boxing in Figure~\ref{fig:hardware}, both platforms coordinated torso, arm, and leg motion with identical feedback weights, validating hardware execution.

\section{Conclusion}
We presented \ours{}, coupling AdaMorph and AdaTrack to connect human-motion retargeting with cross-robot whole-body tracking. AdaMorph synthesizes native references via target-agnostic motion encoding and dual-path robot-conditioned decoding; AdaTrack executes them through morphology-conditioned variable-length attention and a shared action head. Across four simulated platforms, the shared actor achieved 94.00\%/91.96\% seen/unseen success and 49.1/50.9~mm MPKPE with 21.15\% of four specialists' parameters, improved low-data adaptation in held-out-embodiment transfer, and executed squatting, dancing, and boxing on physical G1 and T1 robots, linking reference synthesis, tracking, and control-prior reuse.

\bibliographystyle{IEEEtran}
\bibliography{reference}
\end{document}